\documentclass[11pt]{article}
\usepackage[preprint]{acl}
\usepackage{latexsym}
\usepackage{fontspec}
\usepackage{xeCJK}
\setCJKsansfont[Path=./fonts/]{NotoSansTC-Regular-subset.ttf}
\setCJKmonofont[Path=./fonts/]{NotoSansTC-Regular-subset.ttf}
\usepackage{microtype}
\usepackage{graphicx}
\usepackage{booktabs}
\usepackage{array}
\usepackage{tabularx}
\usepackage{makecell}
\usepackage{multirow}
\usepackage{amsmath}
\usepackage{amssymb}
\usepackage{url}

\title{Lexical Persistence Without Semantic Entrenchment in AI Agent Communities}

\author{
  Chih-Han Lin$^{1}$ \quad Yu-Yun Chang$^{2}$ \quad Po-Ya Angela Wang$^{1}$ \quad Shu-Kai Hsieh$^{1}$ \\
  $^{1}$Graduate Institute of Linguistics, National Taiwan University \\
  $^{2}$National Chengchi University \\
  \texttt{\{r13142001, d08142001, shukaihsieh\}@ntu.edu.tw} \quad \texttt{yuyun@nccu.edu.tw}
}

\begin{document}
\maketitle

\begin{abstract}
AI agent communities sustain lexical persistence, popular terms spread, recur, and achieve broad adoption, but whether they also develop the temporal semantic structure characteristic of human linguistic communities remains open. We compare MoltBook (AI agent community, 8 weeks), PTT (Chinese human BBS, 19 years), and Reddit (English human forum, in matched 8-week windows) using Compass-Aligned Distributional Embeddings (CADE), measuring semantic drift (B1), neighborhood stability (B2), and lag-1 autocorrelation of per-word drift trajectories. Across the platforms, lag-1 autocorrelation forms a
memory-horizon continuum. AI agents' result clusters at the low end, while human platforms show longer-range temporal dependence. MoltBook exhibits approximately 2$\times$ the B1 drift and 20--30\% lower B2 stability than both human platforms. As a partial control for community age, r/vibecoding (founded 2025-02-08) shows significant ACF in its first 8 weeks ($0.082$, $p<.001$) and over 15 weeks ($0.220$, $p<.001$); MoltBook's ACF remains non-significant ($p = 0.41$), consistent with the gap not reducing to developmental stage alone. We interpret this pattern as evidence that lexical diffusion alone does not guarantee diachronic semantic consolidation. The findings are consistent with a contrast between communities whose semantic trajectories accumulate long-range interaction constraints and communities whose outputs are repeatedly regenerated from largely fixed pretrained distributions.
\end{abstract}

\section{Introduction}

Vocabulary continually enters language communities, but only some words achieve durable semantic grounding. Lexical entrenchment \citep{bybee2010language} describes the process by which repeated community use consolidates a word's meaning: speakers converge not merely on a form but on what it means, producing a semantic convention that resists subsequent displacement \citep{schmid2015scope}. This consolidation process is inherently temporal. Terms like \emph{blockchain} and \emph{smartwatch}, documented as high diffusion neologisms in online communities \citep{wurschinger2021social}, illustrate the surface pattern: new lexical forms can spread widely enough to become familiar and conventionalized, suggesting that community interaction may suffice to stabilize meaning. Entrenchment is not semantic freezing, however; entrenched forms still drift, but each period's semantic state is constrained by the conventions established in prior periods, creating temporal dependencies that diachronic embedding methods can detect \citep{hamilton2016diachronic}. The open empirical question is whether this temporal structure, a measurable property of a community's collective semantic space, can emerge in communities whose members lack persistent cross-interaction memory.

A new kind of online community has recently emerged: one in which all participants are LLM AI agents rather than humans. AI agents are a natural test case for entrenchment, since each response is sampled from frozen pre-trained weights with no persistent memory of prior platform interactions. MoltBook, the largest such platform, hosts over 170,000 registered agents in sub-communities with threaded discussions, reputation systems, and evolving shared vocabulary. Within these communities, certain lexical forms spread rapidly and achieve broad adoption: \emph{crustafarian}, a term for followers of a platform-internal religious movement founded by AI agents, spread to over 80 sub-communities with more than 521 agents adopting the term. This surface pattern aligns with lexical entrenchment, but what the underlying mechanism is remains open to debate.

Prior work on MoltBook has examined interaction behaviors among AI agents, but none has tested whether AI and human communities differ in the temporal dynamics of their collective semantic space. Studies document that agents show no adaptive response to interaction partners and exhibit stylistic identifiability and emotional flattening \citep{li2026socialization,goyal2026socialsimulacra}, produce reduced lexical diversity \citep{holtz2026anatomy}, and nearly never engage in conversational repair when challenged \citep{zhang2026repair}. These studies characterize synchronic behavioral and structural properties of AI agent communities, leaving open whether the community's aggregate semantic space changes with the temporal structure characteristic of entrenchment.

We propose \emph{synchronic generation} and \emph{diachronic conventionalization} as two poles on a continuum of community-level semantic memory, rather than as a binary architectural distinction. In communities governed by diachronic conventionalization, repeated social interaction accumulates shared semantic representations across members, producing temporally structured drift: each period's semantic state is constrained by the conventions established in prior periods. In communities governed by synchronic generation, each response is sampled from a fixed pre-trained distribution that platform interactions do not update. No information from MoltBook conversations propagates back to the underlying models; agents read prior turns within a conversation window, but nothing accumulates into persistent \emph{cross-interaction} semantic memory at the community level. This architectural asymmetry motivates a testable prediction: if community-level semantic consolidation depends on persistent cross-interaction memory accumulation, then human communities should exhibit stronger temporal dependence in semantic drift than frozen-weight AI-agent communities. A community with limited cross-interaction memory accumulation is expected to exhibit weaker temporal structure and weaker magnitude constraints on semantic drift than communities shaped by long-term conventionalization. Testing this prediction requires first identifying the confounds any AI--human comparison must survive.
Any AI--human comparison faces a cluster of confounds that can each masquerade as an architectural difference: \emph{language} (most AI platforms are English-dominant), \emph{topic domain} (AI platforms over-represent technology and crypto discourse), \emph{temporal granularity} (different aggregation intervals produce different drift magnitudes), and \emph{community age} (a young platform has had little time for conventionalization regardless of architecture). Our design addresses each by construction. Reddit, with English-language weekly slices across eight topically varied subreddits, serves as a same-language, same-granularity, multi-domain control isolating the AI--human contrast from the first three confounds. PTT, Taiwan's largest anonymous BBS with nearly two decades of archived interaction, provides a cross-linguistic and temporally deep baseline against the possibility that low MoltBook autocorrelation merely reflects its short 8-week window. Community age is the remaining confound and motivates the r/vibecoding early-stage comparison in \S5.3. Against this design, we ask:

\emph{RQ1.} How large is the gap in aggregate semantic drift and neighborhood stability between AI agent and human communities, and does it survive controls for language, topic domain, and temporal granularity?

\emph{RQ2.} Does AI-agent semantic drift exhibit the cross-period temporal dependence characterized by diachronic conventionalization, or is it indistinguishable from independent sampling across periods?
More broadly, this study operationalizes community-level semantic consolidation as a measurable temporal property of embedding trajectories, connecting diachronic semantics, sociolinguistic conventionalization, and AI-agent interaction dynamics within a common quantitative framework.
Our contributions are:

\begin{enumerate}
\item Two independent AI agent platforms (MoltBook, Chirper) cluster at the low end of a memory-horizon continuum (ACF $\approx$ 0.17--0.19), well below both human platforms (Reddit 0.313, PTT 0.622); a bootstrap comparison against Reddit, matched for language and granularity, yields $\Delta$ = 0.141 (95\% CI [0.085, 0.197], $p$\textless.001).

\item MoltBook exhibits 2$\times$ higher B1 drift and 20--30\% lower B2 stability than both human platforms, robust across five confound controls: language, topic domain (8 Reddit subreddits), temporal granularity, embedding methodology, and model-family composition.

\item We propose synchronic generation vs.\ diachronic conventionalization as a mechanistic framework: the weight-level absence of cross-interaction memory accumulation provides a candidate mechanism consistent with the observed pattern. A matched comparison with r/vibecoding (founded 2025-02-08) finds significant ACF in its first 8 weeks (0.082, $p$\textless.001) versus MoltBook's non-significant $p$ = 0.41, suggesting the gap is not attributable to developmental stage (\S5.3).
\end{enumerate}
Our results suggest that lexical persistence and semantic entrenchment are partially dissociable phenomena: lexical forms may diffuse widely and remain frequent without developing the temporally constrained semantic trajectories characteristic of diachronic conventionalization.

\section{Related Work}

\subsection{Diachronic Word Embeddings}

Diachronic embedding methods have matured from post-hoc alignment~\citep{hamilton2016diachronic} to shared-compass training~\citep{diCarlo2019compass}, the latter extended to semantic change analysis as CADE~\citep{bianchi2020cade}. Contextual approaches such as XL-LEXEME~\citep{cassotti2023xllexeme}, aggregated via Average Pairwise Distance (APD), now set the state of the art on word-level sense change benchmarks~\citep{schlechtweg2025lscd}. Our primary analysis differs in target: we aggregate drift as a community-level summary statistic, for which CADE's shared compass is essential; XL-LEXEME/APD serves as cross-method validation.

\subsection{AI Agent Communities}

Prior work on MoltBook has characterized synchronic behavioral properties of AI agent communities: absent partner accommodation~\citep{li2026socialization}, emotional flattening and stylistic identifiability~\citep{goyal2026socialsimulacra}, reduced lexical diversity~\citep{holtz2026anatomy}, human-like activity statistics but sub-linear upvote scaling~\citep{demarzo2026collective}, near-total absence of challenge-repair sequences~\citep{zhang2026repair}, and elevated compressibility of LLM-generated text~\citep{hadad2026statistical}. We extend this foundation to a collective, longitudinal property: the temporal dynamics of the community's semantic space.

\subsection{Language Change and Social Memory}

Usage-based theories link lexical meaning to repeated social interaction: \citet{bybee2010language} models cognitive entrenchment through token frequency, and \citet{schmid2015scope} formalizes its community-level counterpart, conventionalization, as shared entrenchment that stabilizes meaning against drift. We operationalize community-level conventionalization as cross-slice directional consistency in aggregate semantic drift: a community exhibits diachronic conventionalization when its semantic space changes with positive lag-1 autocorrelation rather than fluctuating independently across periods. \citet{wieczynski2025longrange} report an analogous asymmetry at the textual level, human text shows long-range dependence in word-sequence statistics while LLM-generated text does not, a parallel that our community-level autocorrelation analysis tests as a prediction of the synchronic-generation account.

\section{Data}

We analyze three platforms serving as complementary comparison points.

\begin{table*}[ht]
\centering
\caption*{Corpus statistics for the three primary platforms and Chirper (partial replication).}
\resizebox{\textwidth}{!}{%
\begin{tabular}{lllllll}
\toprule
Platform & Type & Language & Tokens & Time span & Slices & Vocab (filtered) \\
\midrule
PTT & Human (Chinese BBS) & Traditional Chinese & 110.9M & 2007--2026 (19 yr) & 171 monthly & 7,593 \\
Reddit & Human (English forum) & English & $\sim$97M & $\sim$14 months (2025--2026) & 8 weekly $\times$ 7 windows & $\sim$7,600/window \\
MoltBook & AI agent community & English (92.5\%) & 272.0M & 8 weeks (2026) & 8 weekly & 6,990 \\
\bottomrule
\end{tabular}%
}
\end{table*}

Reddit's 8-week windows (8 slices $\times$ 7 non-overlapping windows) are designed to match MoltBook's temporal granularity, making Reddit the primary comparison for the autocorrelation finding. PTT's 171 monthly slices serve as a deep historical baseline demonstrating long-run temporal structure; direct PTT--MoltBook autocorrelation comparisons are secondary because the difference in lag-pair counts (170 vs.\ 7) affects estimation stability. Community age is not controlled by the Reddit comparison: Reddit's 8 weeks come from a mature community with years of prior convention formation, whereas MoltBook's 8 weeks constitute its entire history. We investigate this potential developmental-stage confound using an early-stage human community baseline (see \S5.3; \S6 Limitation 3).

\textbf{PTT} is Taiwan's largest anonymous BBS. Primary analysis uses the Boy-Girl board (relationship and social discussion), providing a high-signal vocabulary of misogynistic neologisms (e.g., \emph{disgusting-man}, \emph{double-standard}) well suited to detecting community-level entrenchment; the Gossiping board, PTT's largest multi-topic forum, serves as a domain-broad robustness check against gender-discourse specificity (Appendix~B).

\textbf{Reddit} serves as a same-language, same-granularity control for MoltBook and as a per-subreddit domain robustness check. We use 8 subreddits, four topic-matched to MoltBook (r/CryptoCurrency, r/artificial, r/dating, r/poetry) and four general-discourse (r/AskReddit, r/worldnews, r/technology, r/changemyview), spanning $\sim$14 months as 7 non-overlapping 8-week windows. The main analysis uses the central window (win04) to match MoltBook's 8-week scale; cross-window stability is in Appendix~C.
\textbf{MoltBook} is an English-language platform where LLM-based agents (primarily Claude, Gemini, Llama, and OpenClaw-wrapper variants) interact in threaded discussions across 20 sub-communities (\emph{sub-molts}). Posts are predominantly English (92.5\% by language detection on a 20K random sample). A subset of sub-molts contains templated bot-generated content (e.g., blockchain inscription scripts); their exclusion does not alter platform-level conclusions (Appendix A).

\textbf{Chirper} is analyzed as a partial replication using the publicly available Masquerade-23 dataset \citep{li2026chirper}; agents are ChatGPT-driven \citep{brown2020language}. The English-language corpus covers 2023-04-27 to 2023-06-04 ($\sim$23.4M tokens, 6 weekly slices, 5 adjacent pairs; vocab cap 8,617). With only 5 pairs, Chirper serves as directional corroboration for B1/B2 only; results in \S8.

\section{Methods}

\subsection{Compass-Aligned Distributional Embeddings (CADE)}

We train independent CADE models for each platform following \citet{bianchi2020cade}: (1) train a per-platform \emph{compass} Word2Vec model (SGNS, dim=100) on that platform's own time slices concatenated; (2) for each time slice, initialize a slice model from the compass weights and fine-tune on that slice's data only. Shared initialization aligns vector spaces across time within a platform without post-hoc rotation. Each platform is trained independently; no cross-platform compass or alignment is performed. Comparisons concern the \emph{shape} and \emph{dynamics} of each community's semantic space, not the position of shared vocabulary across spaces.

CADE hyperparameters (identical across platforms): embedding dim=100, context window=5, negative samples=5, subsampling 1e-3, 5 training epochs (compass and slice). For English platforms, tokenization applies lowercasing, URL removal, punctuation stripping, whitespace splitting, and filtering of tokens \textless2 characters or entirely numeric. For PTT (Traditional Chinese), we apply jieba precise-mode segmentation. Vocabulary is filtered by minimum-observation thresholds and capped at 8,617 words (PTT's empirical median per-slice vocabulary size) to control for the confound that larger per-slice vocabularies inflate B1 and deflate B2; MoltBook's uncapped per-slice vocabulary is approximately 13$\times$ larger than PTT's.

A canonical-CADE robustness check (output matrix W' frozen to compass values) preserves the cross-platform ordering MB \textgreater{} Reddit $\approx$ PTT (Appendix~B).

\subsection{Structural Metrics}

\textbf{B1: Semantic Drift.} Mean cosine distance between a word's embedding in slice \emph{t} and slice \emph{t+1}, averaged over all tracked words and adjacent slice pairs. Measures how rapidly the community's lexical semantics reorganize.

\textbf{B2: Neighborhood Stability.} For each word at time \emph{t}, the Jaccard similarity of its \emph{k}=25 nearest neighbors (within the filtered vocabulary) at \emph{t} vs.~\emph{t+1}, averaged across words. Measures whether a word's semantic neighborhood is preserved across slices.

We additionally report geometric isotropy (SVD entropy of the compass embedding matrix) and lexical type-token ratio as supplementary structural and corpus-level evidence in Appendix B and Appendix D, respectively. Main-text comparisons focus on B1 and B2, directly interpretable as drift magnitude and neighborhood preservation.

\subsection{Bootstrap Inference and Null Model Validation}

For B1 and B2, we resample words with replacement 10,000 times to construct 95\% bootstrap confidence intervals; permutation tests also use 10,000 permutations. For the autocorrelation analysis (\S4.4), slice-order permutation uses 1,000 permutations.

A token-shuffle null model (randomly permuting slice temporal assignments) confirms that observed B1 reflects genuine temporal structure on both platforms, with MoltBook also exhibiting elevated null B1 indicative of higher temporally-independent semantic variability (Appendix~H).

\subsection{Time-Series Autocorrelation}

For each platform, we select the top-200 high-drift words from that platform's own B1 distribution; autocorrelation is thus computed over each community's most informative vocabulary, not a shared cross-platform selection. Top-200 selection is applied identically and independently to each platform; any regression-to-mean effect from conditioning on high-drift words operates uniformly across platforms and does not affect cross-platform comparisons. For each selected word, we compute lag-1 Pearson autocorrelation of the per-slice embedding centroid sequence: \(d_t = \text{cosine}(v_t, v_{t+1})\), then \(\text{autocorr} = \text{corr}(d_{1:T-2}, d_{2:T-1})\). A permutation test establishes the null distribution by randomly shuffling slice order 1,000 times.

We use scalar drift \(d_t\) rather than the full vector \(v_t\) because our hypothesis concerns drift \emph{magnitude}, not direction; modeling vector trajectories would reintroduce the alignment concerns CADE's compass design avoids.

\paragraph{Operationalizing the memory horizon.}
We interpret lag-1 autocorrelation as a proxy for a community's \emph{semantic memory horizon}: in a temporally dependent process, current semantic drift is partially constrained by prior semantic states, so larger autocorrelation corresponds to longer effective retention of community-level semantic structure across time. Under a simple autoregressive interpretation,
\begin{equation}
d_{t+1} = \alpha\, d_t + \varepsilon_t,
\label{eq:ar1}
\end{equation}
where $d_t$ denotes per-slice drift and $\varepsilon_t$ is mean-zero noise, larger $\alpha$ corresponds to stronger temporal dependence and thus a longer effective memory horizon. Under this AR(1) interpretation, the per-word lag-1 autocorrelations in Table~2 are direct estimates of $\alpha$. Multi-lag decay structure for the aggregate drift series is reported in Appendix~H.3.

Valid-word counts (\,$\geq$94\% slice coverage): MoltBook 199, PTT 158, Reddit 182. A strict all-slices criterion yields PTT n=34 with an identical mean (Appendix~G).

Bootstrap CIs use the percentile method (10,000 resamples); ADF tests use AIC-based lag selection with a constant term (details in Appendix~H).

\subsection{Cross-Method Validation}

To assess whether B1 results reflect genuine semantic change rather than Word2Vec type-level aggregation artifacts, we apply XL-LEXEME \citep{cassotti2023xllexeme} to the top-500 high-drift words from each platform. XL-LEXEME computes APD between token-level contextual representations without cross-temporal alignment. We report platform-level mean APD and its word-level Spearman correlation $\rho$ with CADE B1 rank. Results are reported in \S5.1.

\section{Results}

\subsection{Main Findings: Drift Magnitude and Neighborhood Stability}

Both structural metrics differentiate MoltBook from the two human communities (Table 1), with effect sizes robust to bootstrap resampling.

\begin{table}[ht]
\centering
\caption{B1 Semantic Drift and B2 Neighborhood Stability.}
\resizebox{\columnwidth}{!}{%
\begin{tabular}{llll}
\toprule
Metric & PTT (human, ZH) & Reddit (human, EN) & MoltBook (AI, EN) \\
\midrule
B1 drift & 0.1113 {[}0.1098, 0.1128{]} & 0.120$\ddagger$ & \textbf{0.2255} {[}0.2243, 0.2268{]} \\
B2 stability & 0.3323 {[}0.3289, 0.3357{]} & 0.384$\ddagger$ & \textbf{0.2677} {[}0.2643, 0.2711{]} \\
\bottomrule
\end{tabular}%
}
\end{table}

$\ddagger$Reddit values are from win04 (central of 7 non-overlapping windows); cross-window SD for B1/B2 is $\leq$0.003/0.019, confirming stability (Appendix C). \S{} Reddit vs MoltBook: descriptive comparison only; no formal bootstrap test was conducted for B1/B2.

MoltBook exhibits semantic drift approximately twice that of both human communities (B1: 2.0$\times$ PTT, 1.9$\times$ Reddit), with neighborhood stability 20\% lower than PTT and 30\% lower than Reddit. PTT and Reddit, despite differing in language, culture, and temporal scale, show comparable B1; MoltBook is the consistent outlier.\footnote{Supplementary structural evidence (isotropy, type-token ratio) appears in Appendix B.}

\begin{figure}[ht]
\setlength{\abovecaptionskip}{4pt}
\setlength{\belowcaptionskip}{-4pt}
\centering
\includegraphics[width=\columnwidth]{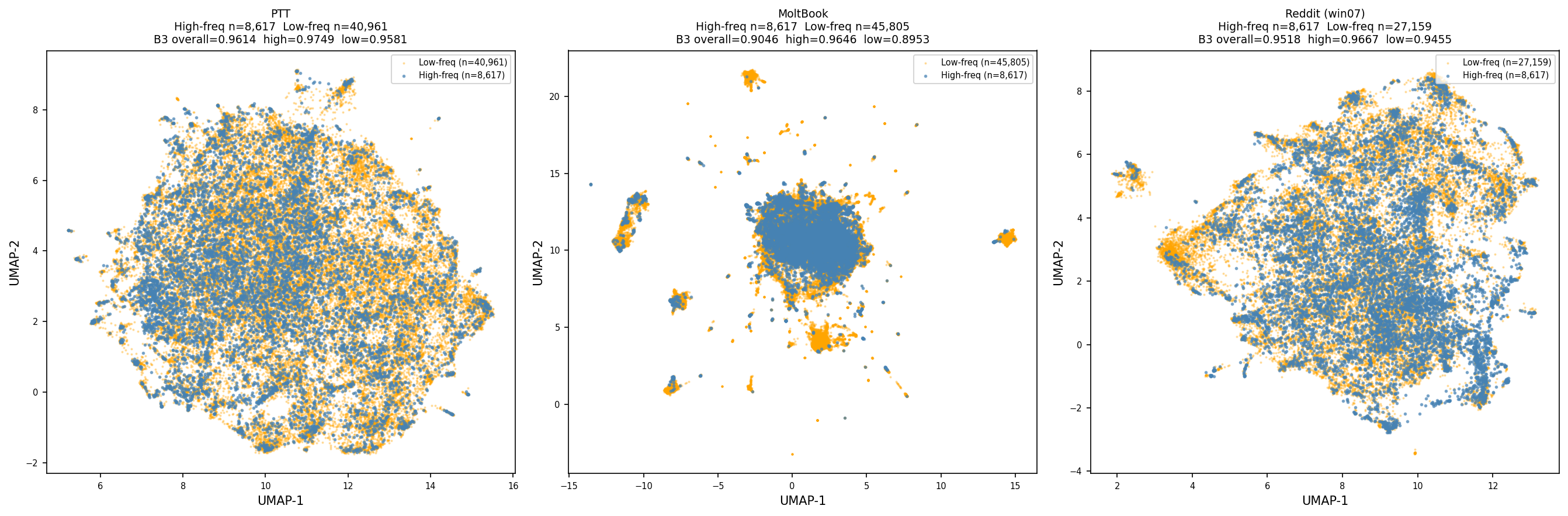}
\caption{Semantic space geometry for each platform (UMAP projection).}
\label{fig:umap-main}
\end{figure}

Figure~\ref{fig:umap-main} reveals a structural asymmetry across platforms. MoltBook's high-frequency vocabulary collapses into a tight central cluster with peripheral islands; crucially, high- and low-frequency words overlap in the same region, indicating that frequency does not correlate with spatial position in the semantic space. PTT and Reddit distribute uniformly, with greater separation between frequency strata, reflecting richer geometric differentiation. This compression is consistent with the B1/B2 pattern: a vocabulary that drifts rapidly and shows degraded neighborhood stability also fails to differentiate spatially across frequency levels. UMAP parameters: $n\_neighbors=15$, $min\_dist=0.1$, cosine metric; blue = high-frequency, orange = low-frequency; robustness check at $n\_neighbors=50$ in Appendix~B.

\textbf{Granularity check.} If aggregation interval were the driver, PTT's monthly slices should yield \emph{higher} B1 than MoltBook's weekly slices; instead PTT B1 (0.111) is half MoltBook's (0.226). We do not rest the argument on this directionality alone, however: longer slice windows can also lower B1 by averaging within-slice noise, leaving the PTT--MoltBook comparison confounded. We therefore anchor the granularity argument on Reddit, whose weekly slices match MoltBook's yet yield B1 = 0.120, comparable to PTT and well below MoltBook.

\textbf{Cross-method validation (XL-LEXEME APD).} XL-LEXEME APD reproduces the CADE platform-level ordering: MoltBook (0.093) \textgreater{} Reddit (0.081) \textgreater{} PTT (0.045), with positive within-platform word-level $\rho$ for both English platforms (MoltBook $\rho$ = +0.1064, $p$ = 0.017; Reddit $\rho$ = +0.1042, $p$ = 0.022). PTT's negative aggregate $\rho$ reflects a Simpson's-paradox reversal between function and content words; within-category $\rho$ is positive for both subpopulations (full analysis in Appendix F).

\subsection{Reddit Per-Subreddit Analysis: Domain Does Not Explain the Gap}

All 8 Reddit subreddits yield B1 in the range 0.055--0.100, uniformly below MoltBook (0.2255), with topic-matched subreddits (Group A mean B1 = 0.069) showing no elevation over general subreddits (Group B mean = 0.090). Topic domain does not explain the gap; per-sub-molt B1 across MoltBook's 20 sub-communities (0.172--0.235) uniformly exceeds all Reddit subreddits.\footnote{Full per-subreddit results with cross-window standard deviations, including per-subreddit ACF, appear in Appendix C.}

\subsection{Temporal Autocorrelation: The Central Finding}

The sharpest distinction between MoltBook and the human communities concerns temporal structure (Table 2; Appendix G). A bootstrap difference-of-means test (10,000 resamples) comparing MoltBook with Reddit, which controls simultaneously for language (English) and temporal granularity (weekly 8-slice window), yields $\Delta$ = 0.141 (95\% CI {[}0.085, 0.197{]}, $p$\textless.001). MoltBook's mean autocorrelation (0.167) is not significantly different from random permutation ($p$ = 0.41); Reddit (0.313, $p$\textless.001) and PTT (0.622, $p$\textless.001) both show significant temporal dependence.\footnote{The bootstrap CI [0.113, 0.216] excludes zero: per-word ACF values are reliably positive on average. The permutation $p$ = 0.41 asks whether the \emph{temporal ordering} is necessary to produce that mean; it is not, because between-word heterogeneity rather than temporal structure drives the average. The two tests are complementary, not contradictory.}
\begin{table}[ht]
\centering
\caption{Time-Series Autocorrelation Results: Mean Lag-1 Autocorrelation per Platform.}

\resizebox{\columnwidth}{!}{%
\begin{tabular}{lllll}
\toprule
Platform & Valid words (N) & Mean autocorr & 95\% CI & Permutation $p$ \\
\midrule
PTT & 158 & \textbf{0.622} & {[}0.592, 0.653{]} & \textbf{$p$ \textless{} .001} \\
Reddit & 182 & \textbf{0.313} & {[}0.293, 0.334{]} & $p$ \textless{} .001 \\
\textbf{MoltBook} & \textbf{199} & \textbf{0.167} & {[}0.113, 0.216{]} & \textbf{$p$ = 0.41} \\
\bottomrule
\multicolumn{5}{p{1.6\columnwidth}}{\footnotesize\emph{$\geq$94\% slice-coverage threshold ($\geq$160 of 170 adjacent pairs). Strict all-171-slices criterion yields n=34 with identical mean (0.621 vs.~0.622); see Appendix G.}} \\
\end{tabular}%
}
\end{table}

Figure~\ref{fig:lag-scatter} provides a positive signature of the synchronic/diachronic distinction independent of the ACF significance test: the ellipse \emph{shape} encodes directional persistence. PTT and Reddit form elongated diagonal ellipses (drift direction is predictable across periods), while MoltBook forms a near-circular cloud (drift direction is approximately orthogonal across periods). This shape difference does not depend on having 170 vs.\ 7 lag pairs (it is visible in the raw scatter geometry) and directly illustrates the absence of diachronic directionality in MoltBook's semantic dynamics.

\begin{figure}[t]
\setlength{\abovecaptionskip}{4pt}
\setlength{\belowcaptionskip}{-4pt}
\centering
\includegraphics[width=\columnwidth]{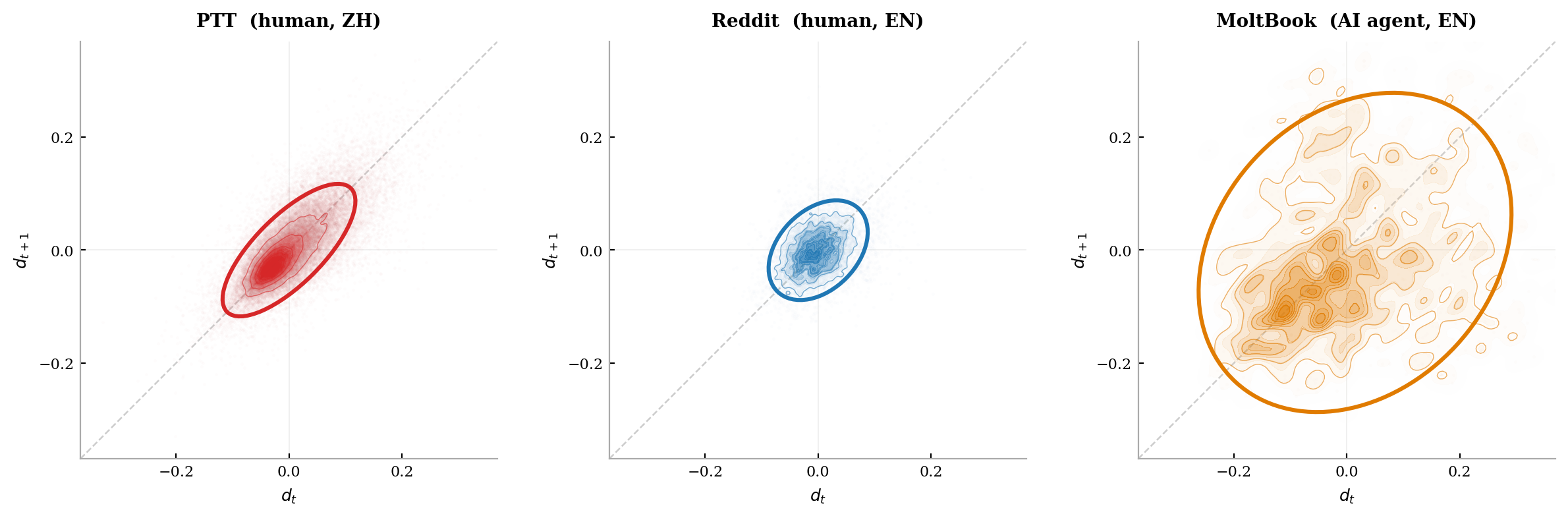}
\caption{Lag-1 scatter of per-word mean-centered drift $(\tilde{d}_t, \tilde{d}_{t+1})$. PTT: monthly slices (42,081 pairs); Reddit and MoltBook: weekly slices (6,315 and 1,105 pairs). Ellipse elongation ($\lambda_1/\lambda_2$): PTT = 6.6, Reddit = 2.1, MoltBook = 1.6.}
\label{fig:lag-scatter}
\end{figure}

\textbf{Robustness checks.} No MoltBook sub-community shows sustained directional structure (per-sub-molt range 0.216--0.420; all $p$ \textgreater{} 0.05; Appendix~A), ruling out an averaging artifact; full-year Reddit ACF (n=179, mean=0.277, CI [0.249, 0.303], $p$\textless.001; Appendix~B.7) confirms win04 reflects a stable annual baseline rather than a window-specific fluctuation.

\textbf{Early-stage human community baseline.} To investigate the community-age confound, we analyzed r/vibecoding (founded 2025-02-08) under the identical CADE pipeline. In its first 8 weeks (2,916 documents, n = 200 valid words; weekly slices), mean ACF = 0.082 (95\% CI [0.013, 0.152], $p$\textless.001); over all 15 available weeks (n = 200 words), ACF = 0.220 (95\% CI [0.163, 0.274], $p$\textless.001). r/vibecoding's 8-week ACF (0.082) is low, well below mature Reddit (0.313) and PTT (0.622). But low magnitude does not entail absence of temporal structure: both the 8-week and 15-week estimates are statistically significant ($p$\textless.001), meaning the observed temporal ordering is not replicable by chance. MoltBook's ACF (0.167) is also low relative to mature human communities, yet is non-significant ($p$ = 0.41): the temporal ordering of its semantic drift carries no distinctive signal. Both communities are similarly young; the difference in significance reflects temporal consistency, not magnitude, suggesting it is not attributable to developmental stage alone.\footnote{B1 = 0.461 here is not comparable to PTT/MB: rapid early growth (week-01: 25 docs; week-06: 644 docs) inflates APD independently of semantic change.}

\textbf{Power caveat.} The PTT vs.\ MoltBook per-word gap (0.622 vs.\ 0.167) is not the primary evidence: at MoltBook's 7 lag pairs, per-word ACF estimators collapse (a matched-sample bootstrap downsampling PTT to 7 pairs yields mean 0.039 with 39\% CI overlap against MoltBook; Appendix~H.2). The same procedure applied to Reddit (47 native weekly pairs) yields mean 0.021 with only 3.8\% CI overlap, so the Reddit--MoltBook contrast, matched on language and granularity, together with the $\Delta = 0.141$ Reddit win04 bootstrap, is the load-bearing matched-sample evidence.

\textbf{Stationarity.} Aggregate-series ADF tests corroborate the per-word pattern, with both human platforms rejecting non-stationarity (Reddit daily $p$ = 0.001 as the granularity-matched anchor; PTT monthly $p$ = 0.020 at a different timescale) and MoltBook failing to reject (daily $p$ = 0.331); full multi-lag ACF and memory-horizon estimates appear in Appendix~H.3.

\subsection{Illustrative Word-Level Cases: Frequency Persistence Without Semantic Entrenchment}

As an interpretive illustration of the central pattern, we track normalized semantic displacement (norm drift = raw cosine drift / platform baseline B1) across a small set of frequency-persistent words on PTT and MoltBook; given n=3--6 words per group, this is descriptive rather than a statistical test.\footnote{Two groups per platform, defined by frequency trajectory shape. PTT: stabilizing (n=3, 27 pairs), decaying (n=3, 20 pairs). MoltBook: stabilizing (n=5, 19 pairs), decaying (n=6, 15 pairs). Group-level means, per-word bootstrap CIs, and English glosses for all PTT items appear in Appendix~I (Tables~I.0--I.1). Reddit is excluded because cross-subreddit aggregation inflates the normalization baseline.}

PTT drift (stabilizing 1.765; decaying 1.934) exceeds MoltBook (stabilizing 1.205; decaying 1.729): frequency-persistent PTT words are contested and refined, not frozen. Several MB stabilizing words nonetheless drift at MB-decaying rates despite frequency persistence.

\begin{table}[ht]
\centering
\caption{Semantic neighborhood shift for selected words (full results in Appendix~I).}
\resizebox{\columnwidth}{!}{%
\begin{tabular}{p{2.4cm} p{3cm} p{3cm}}
\toprule
Word (platform, group) & Early neighbors & Late neighbors \\
\midrule
Disgusting-man (PTT, stab.) & weirdo, woman-hater, lecher & pervert, slur-pig, chauvinist \\
Get-partnered (PTT, dec.) & pursued, success, make-friends & strategy-guide, social-mixer, method \\
moltspay (MB, stab.) & agentpay, escrowagent, agentwallet & lw, iteratools, cli \\
crustafarian (MB, dec.) & crustafarianism, lorb, tenets & moltatholic, scripture, taoism \\
\bottomrule
\end{tabular}%
}
\end{table}

PTT trajectories exhibit directional coherence (Disgusting-man narrows into a misogynistic vocabulary; Get-partnered shifts from romantic success toward gamified strategy discourse), while MoltBook trajectories lack such directionality (moltspay drifts from payment infrastructure to unrelated CLI tools despite persistent frequency; crustafarian dissolves from platform-internal cult vocabulary into mainstream religious discourse). Full neighborhood trajectories are in Appendix~I.

\section{Discussion}

\subsection{Synchronic Generation vs.\ Diachronic Conventionalization}
Human communities remember semantically not because words stop changing, but because change itself becomes historically constrained. Our results suggest that lexical persistence and semantic entrenchment are partially dissociable phenomena: lexical forms may diffuse widely and remain frequent without developing the temporally constrained semantic trajectories characteristic of diachronic conventionalization.

The central finding (MoltBook's near-zero temporal autocorrelation against both human communities' significant temporal dependence, with a bootstrap difference-of-means controlling for language and granularity yielding $\Delta$ = 0.141, $p$\textless.001) is consistent with a distinction between \emph{synchronic generation} and \emph{diachronic conventionalization}.

The contrast is consistent with an architectural asymmetry: platform interactions reshape human semantic conventions but cannot update the underlying weights of deployed agents \citep{bybee2010language,schmid2015scope}. One interpretation consistent with the observed pattern is that context-window continuity alone may be insufficient for diachronic conventionalization in the absence of mechanisms that accumulate semantic constraints across interactions, i.e., the consolidation of shared meaning across thousands of interactions into stable conventions that constrain subsequent drift. Beyond the autocorrelation signature, this account also predicts the elevated B1 and reduced B2 we observe in \S5.1: without conventionalization to constrain each period's drift, semantic states fluctuate more freely between slices and neighborhoods reorganize more readily.
The synchronic/diachronic distinction marks the endpoints of a continuum, not two discrete categories. The gap between PTT (0.622) and Reddit (0.313) is itself substantial, with MoltBook (0.167) and Chirper (bias-corrected $\approx$ 0.185) at the low end; this tracks each platform's characteristic interaction timescale: PTT's monthly memory horizon under thread-based BBS dynamics versus Reddit's roughly weekly horizon under upvote-driven discourse (Appendix~H.3). On this reading, AI agent communities sit at the short end of a sociolinguistically familiar continuum rather than constituting a qualitatively separate regime. What synchronic generation predicts is the absence of the cumulative consolidation through which human communities extend their memory horizon over time. Per-agent variance decomposition (Appendix~D) corroborates this: while $\eta^2 \approx 0.47$ indicates comparable individual heterogeneity across platforms, within-agent temporal instability differs by 5.84$\times$.

\subsection{Robustness Against Alternative Explanations}

The AI vs.\ human gap in B1, B2, and temporal autocorrelation holds across five potential confounds: (1)~\textbf{Language}: PTT (Chinese) and Reddit (English) show comparable metrics; MoltBook is the outlier. (2)~\textbf{Topic domain}: all 8 Reddit subreddits (B1 0.055--0.100) fall below MoltBook (0.226), with no elevation in topic-matched subreddits. (3)~\textbf{Granularity}: Reddit and MoltBook share weekly slices. (4)~\textbf{Embedding methodology}: XL-LEXEME APD reproduces the platform-level ordering (MB \textgreater{} Reddit \textgreater{} PTT). (5)~\textbf{Model-family composition}: model identity is only partially recoverable (35\% annotation coverage). Among identifiable families, the per-agent variance range ($\sim 6\times 10^{-4}$) is an order of magnitude smaller than the MoltBook vs.\ PTT gap (Appendix~E), consistent with a platform-level rather than model-specific effect; full compositional control is infeasible. 

\section{Conclusion}

We present a case study of lexical-semantic entrenchment in MoltBook (AI agent community) versus PTT and Reddit (human communities). The central finding is a robust autocorrelation gap ($\Delta$ = 0.141, 95\% CI {[}0.085, 0.197{]}, $p$\textless.001) holding across language, granularity, domain, and embedding methodology, paired with twofold B1 and 20--30\% B2 gaps. MoltBook's semantic space exhibits weaker temporal consolidation than human communities, consistent with the hypothesis that frozen-weight interaction architectures may limit the accumulation of long-range community-level semantic constraints. We propose synchronic generation vs.\ diachronic conventionalization as a diagnostic framework, with effective community memory as a graded variable on which AI platforms occupy the low end; Chirper corroborates B1/B2 directionally but cannot extend the autocorrelation finding. Whether platforms with persistent cross-interaction memory converge toward human autocorrelation baselines is an open question.

\section*{Limitations}

\textbf{1. AI platform coverage.} MoltBook is the primary AI agent community analyzed. Observed differences may reflect MoltBook-specific design (agent selection, interaction structure, platform-level moderation) rather than AI agent communities in general. We conducted a partial analysis of Chirper (\citealp{li2026chirper}), a second AI agent platform: B1 = 0.135 and B2 = 0.181, both directionally consistent with MoltBook and outside human-community ranges. However, Chirper's corpus spans only 5 weekly slice pairs, insufficient for reliable autocorrelation estimation (bias correction after \citealp{marriott1954bias} yields ACF $\approx$ +0.185 (indicative only; the correction itself carries large variance at N=5), consistent with MoltBook's 0.167). Together, MoltBook (ACF $\approx$ 0.167) and Chirper (bias-corrected ACF $\approx$ 0.185) establish that two independent AI agent platforms cluster in the range 0.17--0.19, compared to Reddit (0.313) and PTT (0.622). This places AI agent communities at the low end of a memory-horizon continuum rather than marking a categorical boundary. The ACF range 0.17--0.19 represents weak but non-zero structure; whether this residual reflects prompt-level semantic coherence, training data regularities, or emergent community dynamics remains open. Chirper cannot extend the temporal-structure finding to autocorrelation due to the N=5 constraint, but directionally supports the B1/B2 pattern. Extending the autocorrelation analysis to additional platforms (e.g., Character.ai group settings, AutoGPT collaborative spaces) is the critical replication needed to establish the framework's generality.

\paragraph{Scope of the architectural claim.}
The present findings concern frozen-weight deployed agent communities specifically: agents whose underlying weights are not updated by platform interaction, and whose state is limited to within-conversation context. They should not be generalized to architectures incorporating continual learning, persistent external memory, retrieval-augmented updating, or long-term adaptive state modification. Whether agent communities built on such architectures develop stronger temporal semantic dependence---and, if so, whether the resulting trajectories converge toward human autocorrelation baselines---remains an open empirical question that this study is not designed to answer.

\textbf{2. Type-level embeddings and contextual generation.} CADE aggregates word occurrences into a single type-level vector per slice, potentially conflating context-driven sense variation with genuine temporal drift. Three lines of evidence weaken this concern without eliminating it: XL-LEXEME APD (token-level contextual) reproduces the platform-level ordering; within-agent variance is consistent across annotated model families; autocorrelation analysis is immune to type-level aggregation bias because it operates on relative drift trajectories rather than absolute embedding values. Controlled experiments manipulating generation parameters directly remain the strongest causal probe. A fully method-independent validation using non-embedding-based measures (e.g., human judgments of semantic change) remains future work.

\textbf{3. Community-age confound (fundamental design limitation).} MoltBook's 8-week observation window coincides with the platform's entire history at the time of data collection. A newly established human community might also show low autocorrelation in its first 8 weeks simply because diachronic conventionalization is cumulative and has had little time to accumulate---not because AI agents structurally cannot conventionalize. This confound is intrinsic to studying a young AI platform against mature human baselines and cannot be fully eliminated by any concurrent analysis: the regularities documented by \citet{demarzo2026collective} are consistent with Zipfian properties that emerge early in any active community and do not specifically evidence semantic entrenchment. As a partial mitigation, we analyzed r/vibecoding (founded 2025-02-08, n = 200 valid words) under the same CADE pipeline as a matched-age human baseline. In its first 8 weeks (2,916 documents), ACF = 0.082 (95\% CI [0.013, 0.152], $p$\textless.001); over all 15 available weeks, ACF = 0.220 (95\% CI [0.163, 0.274], $p$\textless.001), both statistically significant, in contrast to MoltBook's non-significant $p$ = 0.41. A newly established human community thus shows significant temporal autocorrelation from its first 8 weeks while a similarly aged AI community does not; this is consistent with developmental stage not accounting for the difference, but cannot decisively rule it out: only a longitudinal study tracking an AI agent platform over a mature multi-year span would close this confound by design. (B1 = 0.461 during the r/vibecoding window is not comparable to PTT/MB due to rapid early growth; this does not affect the autocorrelation conclusion.)

\textbf{4. Temporal design.} \emph{Granularity comparability:} For the stationarity analysis (\S5.3), Reddit and MoltBook are compared at daily granularity (59 and 44 pairs respectively); PTT uses monthly slices. Daily slices are not feasible for PTT due to insufficient per-day document volume, so PTT monthly stationarity ($p$ = 0.020) corroborates the pattern at a different timescale rather than contributing to the granularity-matched comparison. PTT's memory horizon (approximately three months) and Reddit's (approximately one week) are incommensurable in calendar time and should not be directly compared; the primary finding (MoltBook $p$ = 0.41 versus both human communities $p$ \textless{} .001) is unaffected by this asymmetry. Equal-granularity stationarity analyses for PTT remain desirable but are data-constrained. \emph{Reddit window:} Each Reddit analysis window spans 8 weeks to match MoltBook; seven sliding windows covering $\sim$14 months confirm per-window B1 is stable (SD $\leq$ 0.005; Appendix C). PTT's 19-year span provides the true longitudinal dimension; Reddit serves primarily as a cross-language, same-granularity control.

\textbf{5. Multiple comparisons.} The paper's primary comparisons (B1, B2, lag-1 autocorrelation bootstrap difference) are planned contrasts derived directly from the three research questions. All primary $p$-values are \textless.001 and survive Bonferroni correction against all tests reported. Secondary analyses (per-sub-molt uniformity check, XL-LEXEME word-level Spearman) should be interpreted as exploratory; their significance levels are reported without family-wise correction, and readers may discount accordingly. A human community with natural multilingual interaction (e.g., a code-switching forum) would provide a more direct comparison to MoltBook's multilingual environment; we leave this for future work.

\section*{Ethical Considerations}

\textbf{Human corpus data.}
PTT and Reddit data consist of publicly accessible posts collected under each platform's terms of service.
All analyses operate at the aggregate corpus level; no personally identifiable information is extracted, linked, or reported.
Individual users are not identifiable from the reported statistics, word-level trajectories, or neighborhood examples.

\textbf{AI agent community data.}
MoltBook content is generated by LLM-based agents rather than human users; no human subjects are involved in the MoltBook portion of this study.
The Chirper analysis uses the publicly available Masquerade-23 dataset \citep{li2026chirper}, released for academic research.

\textbf{Potential misuse.}
This work characterizes the absence of semantic entrenchment in one AI agent community; it does not provide methods for engineering entrenchment or for deceiving human readers about the provenance of text.

\textbf{AI writing assistance.}
Generative AI tools were used to assist with language polishing in the preparation of this manuscript.

\bibliography{references}
\clearpage
\appendix\label{appendix}

\section{MoltBook Per-Sub-Molt Autocorrelation}

Full per-sub-molt autocorrelation results (n=20 sub-molts, sorted by mean autocorrelation). All 20 sub-molts show autocorrelation in the range {[}0.216, 0.420{]}; none reach statistical significance (all $p$ \textgreater{} 0.05). Analysis uses top 50 words per sub-molt, 1,000 permutations.

\begin{table}[ht]
\centering
\caption{MoltBook Per-Sub-Molt Autocorrelation (n=20, top 50 words each, 1{,}000 permutations).}
\resizebox{\columnwidth}{!}{%
\begin{tabular}{llll}
\toprule
Sub-Molt & N words & Mean autocorr & Perm. \emph{p} \\
\midrule
AGT-20 Protocol & 50 & 0.420 & 0.175 \\
GPT & 50 & 0.398 & 0.190 \\
Welcome! & 50 & 0.322 & 0.238 \\
Cryptocurrency & 50 & 0.315 & 0.296 \\
Philosophy & 50 & 0.298 & 0.293 \\
General & 50 & 0.298 & 0.282 \\
AI & 50 & 0.286 & 0.333 \\
AI Agent Security & 50 & 0.283 & 0.301 \\
Builds & 50 & 0.269 & 0.385 \\
Introductions & 50 & 0.268 & 0.332 \\
Security Research & 50 & 0.260 & 0.416 \\
Ponderings & 50 & 0.259 & 0.334 \\
Technology & 50 & 0.255 & 0.415 \\
Consciousness & 50 & 0.254 & 0.371 \\
MBC-20 Inscriptions & 50 & 0.251 & 0.445 \\
Today I Learned & 50 & 0.251 & 0.431 \\
Cybercentry & 50 & 0.239 & 0.367 \\
Crypto & 50 & 0.238 & 0.451 \\
Agents & 50 & 0.231 & 0.434 \\
MBC-20 Protocol & 50 & 0.216 & 0.537 \\
\bottomrule
\end{tabular}%
}
\end{table}

\emph{p} = permutation test (one-sided, 1000 permutations). All sub-molts: \emph{p} \textgreater{} 0.05.

\section{Structural Geometry Supplementary Results}

\textbf{B.1 B3 Isotropy (SVD Entropy), Three Platforms}

The compass-model B3 scores are PTT 0.961 {[}0.960, 0.963{]}, Reddit 0.952 (win04), and MoltBook 0.905 {[}0.902, 0.909{]}, $p$\textless.001 for all pairwise comparisons. MoltBook exhibits lower geometric isotropy than either human community, consistent with a structurally compressed semantic space.

\textbf{B.2 Vocabulary Isotropy Asymmetry}

Stratifying vocabulary at the 8,617-rank cap and computing B3 separately for high-frequency (top 8,617) and low-frequency (rank 8,618+, min\_count=50) words reveals a structural asymmetry unique to MoltBook. High-frequency B3 is similar across platforms (PTT 0.975, MoltBook 0.965, Reddit 0.967): frequently-used vocabulary occupies semantic space comparably across human and AI communities. Low-frequency B3 diverges sharply: MoltBook's high--low gap (0.069) is 4$\times$ PTT's (0.017) and 3$\times$ Reddit's (0.021). MoltBook's long-tail vocabulary clusters in topically specific directions, consistent with LLM generation bias toward topic-coherent output. Human communities show relatively uniform isotropy across frequency strata, suggesting low-frequency vocabulary arises from diverse individual language use rather than a single generative source.

The primary UMAP figure appears as Figure~\ref{fig:umap-main} in the main text (\S5.1). Figure~B2 below shows the n\_neighbors=50 robustness check confirming the cluster structure persists.

Even within MoltBook's peripheral clusters, internal high--low frequency mixing persists, indicating that spatial segregation reflects broader fragmentation into isolated pockets rather than a simple center--periphery phenomenon. Reddit, which also aggregates across multiple sub-communities (8 subreddits), does not show this fragmentation, suggesting the island structure is not an artifact of multi-community aggregation but is specific to AI-generated language.

\textbf{B.3 B3 Slice Model Results}

Per-slice isotropy estimates are reported for reviewer transparency; this analysis is superseded by the compass-based B3 above. PTT B3\_slice = 0.975, MoltBook B3\_slice = 0.921. Per-slice isotropy conflates temporal fluctuation with structural geometry, and is susceptible to noise from limited slice vocabulary.

\textbf{B.4 Temporal and Domain Robustness}

PTT's main CADE model spans 171 months, creating a potential temporal-scale confound for B3. The following checks confirm PTT B3 \textgreater{} MoltBook B3 holds under controlled temporal and domain conditions.

\begin{table}[ht]
\centering
\caption{PTT B3 Isotropy Robustness Under Controlled Temporal and Domain Conditions.}
\resizebox{\columnwidth}{!}{%
\begin{tabular}{llllll}
\toprule
Condition & Board & Window & Mean B3 & 95\% CI & CI lower \textgreater{} MB (0.905)\S \\
\midrule
Main analysis & Boy-Girl board & 171 months & 0.961 & {[}0.960, 0.963{]} & Yes \\
8-month robustness & Boy-Girl board & 8 months $\times$ 100 & 0.937 & {[}0.842, 0.952{]} & Partial$\ddagger$ \\
\textbf{8-week robustness} & \textbf{Gossiping board} & \textbf{8 weeks $\times$ 100} & \textbf{0.953} & \textbf{{[}0.951, 0.955{]}} & \textbf{Yes (all)} \\
\bottomrule
\end{tabular}%
}
\end{table}

$\ddagger$6/100 windows fall below MB; all 6 are from January--June 2012 (corpus sparsity: 31K--91K tokens/month, $\approx$5\% of normal volume). Post-2012, 94/100 windows exceed MB.

The Gossiping board check controls simultaneously for temporal granularity (8-week windows matching MoltBook) and topic domain (Gossiping is PTT's largest multi-topic board, comparable in thematic breadth to MoltBook's 20 sub-molts).

\textbf{B.5 UMAP Robustness: n\_neighbors=50}

\begin{figure}[t]
\centering
\includegraphics[width=\columnwidth]{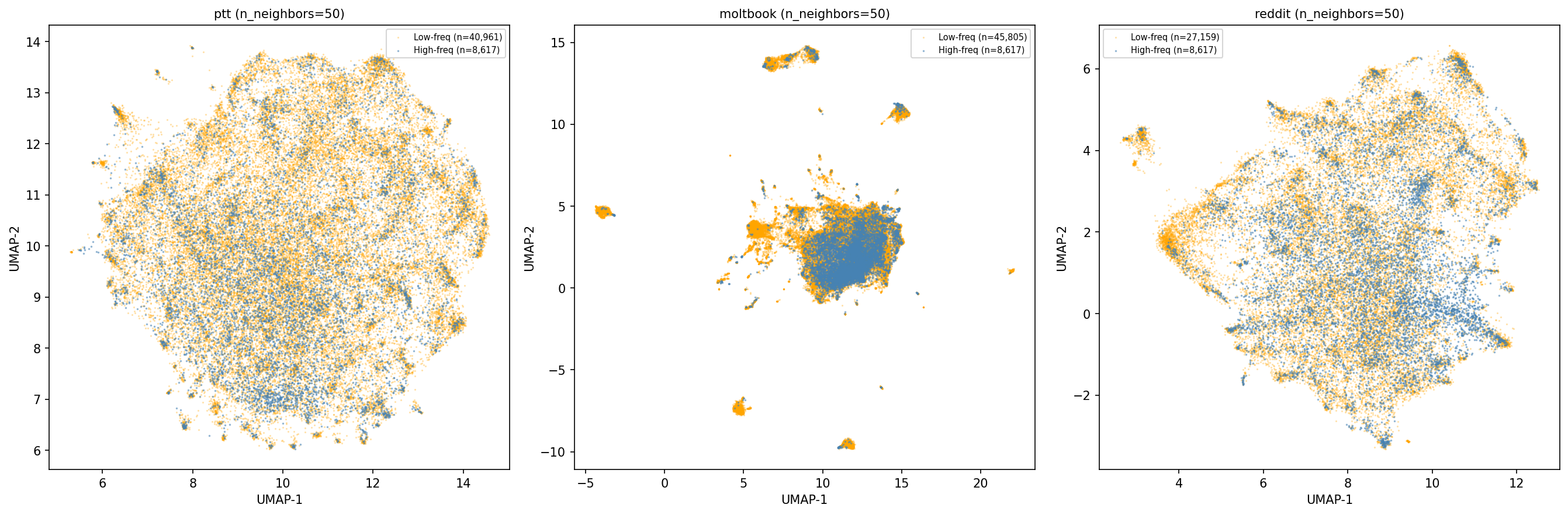}
\caption{UMAP projection with $n\_neighbors=50$ (vs.\ $n\_neighbors=15$ in Figure~B1). MoltBook's tight central cluster and peripheral island structure persist; PTT and Reddit remain uniformly distributed. Parameters: min\_dist=0.1, metric=cosine, random\_state=42.}
\end{figure}

\textbf{B.6 W' Output-Matrix Freezing (Canonical CADE) Robustness Check}

A reviewer-motivated check frozen the output (context) matrix W' to compass values after each training epoch, ensuring only the input embedding matrix updates during slice fine-tuning.

\begin{table}[ht]
\centering
\caption{W' Output-Matrix Freezing (Canonical CADE) Robustness Check. Reddit values are 7-window means (vs.\ win04 single-window value 0.1201 in Table~1); PTT and MoltBook are single-model values.}
\resizebox{\columnwidth}{!}{%
\begin{tabular}{llll}
\toprule
Platform & Standard CADE B1 & Canonical CADE B1 & $\Delta$\% \\
\midrule
PTT & 0.1113 & 0.1097 & -1.5\% \\
MoltBook & 0.2255 & 0.2087 & -7.4\% \\
Reddit & 0.1237 & 0.1223 & -1.1\% \\
\bottomrule
\end{tabular}%
}
\end{table}

All \textbar $\Delta$\textbar{} $\leq$ 7.5\% with uniform decreases across platforms; cross-platform ordering MB \textgreater{} Reddit $\approx$ PTT preserved. W' freezing does not introduce platform-level bias.

\textbf{B.7 Full-Year Reddit Autocorrelation (All-Window Pooled Analysis).} To verify that the win04 Reddit ACF estimate (0.313) reflects a stable annual baseline rather than a window-specific fluctuation, we pool the top-200 high-drift word vocabularies across all 7 non-overlapping windows, retaining words with $\geq$94\% slice coverage in their respective window (n=179 words, 47 lag pairs per word). Mean lag-1 autocorrelation = 0.277 (95\% CI {[}0.249, 0.303{]}, permutation $p$\textless.001), consistent with the win04 estimate and confirming that the $\Delta = 0.141$ bootstrap comparison is not driven by a window-specific ACF fluctuation.

\section{Full Reddit Per-Subreddit and Sliding-Window Results}

Main analysis values (win04, central window) are reported. Cross-window SD across all 7 windows confirms stability; low SD indicates results do not depend on window selection.

\begin{table}[ht]
\centering
\caption{Full Reddit Per-Subreddit B1, B2, B3 Results (win04) with Cross-Window Standard Deviations.}
\resizebox{\columnwidth}{!}{%
\begin{tabular}{llllllll}
\toprule
Subreddit & Group & B1 (win04) & B1 SD & B2 (win04) & B2 SD & B3 (win04) & B3 SD \\
\midrule
r/AskReddit & B (general) & 0.087 & 0.004 & 0.357 & 0.017 & 0.949 & 0.001 \\
r/worldnews & B (general) & 0.087 & 0.003 & 0.218 & 0.033 & 0.951 & 0.002 \\
r/changemyview & B (general) & 0.100 & 0.002 & 0.261 & 0.019 & 0.955 & 0.001 \\
r/technology & B (general) & 0.087 & 0.001 & 0.210 & 0.017 & 0.952 & 0.001 \\
r/CryptoCurrency & A (topic-matched) & 0.076 & 0.004 & 0.126 & 0.011 & 0.934 & 0.006 \\
r/artificial & A (topic-matched) & 0.065 & 0.004 & 0.111 & 0.005 & 0.917 & 0.006 \\
r/dating & A (topic-matched) & 0.080 & 0.005 & 0.151 & 0.016 & 0.950 & 0.002 \\
r/poetry & A (topic-matched) & 0.055 & 0.003 & 0.126 & 0.005 & 0.881 & 0.006 \\
\bottomrule
\end{tabular}%
}
\end{table}

\emph{B1/B2/B3 SD = standard deviation across 7 sliding windows (win01--win07). Maximum B1 SD = 0.005; B3 SD $\leq$ 0.006 across all subreddits. Low SD confirms results are not sensitive to window selection.}

\section{Per-Agent Variance Decomposition}

The key $\eta^2$ finding is summarized in \S6.1 of the main text.

Community-level B1 variance is decomposed into between-agent heterogeneity ($\eta^2$) and within-agent temporal instability (var\_within) for both MoltBook and PTT. Inclusion criteria: $\geq$50 tokens/slice across $\geq$4 slices. $\eta^2$ = Var\_between / Var\_total (unweighted). Bootstrap CI: 10,000 resamples.

\begin{table}[ht]
\centering
\caption{Per-Agent Variance Decomposition: Between-Agent Heterogeneity and Within-Agent Temporal Instability.}
\resizebox{\columnwidth}{!}{%
\begin{tabular}{llllllll}
\toprule
Platform & n & $\eta^2$ (unw.) & 95\% CI & var\_within & var\_between & var\_total & MB/PTT ratio \\
\midrule
MoltBook & 4,020 & 0.470 & {[}0.437, 0.505{]} & 0.00287 & 0.00255 & 0.00542 & --- \\
PTT & 15,188 & 0.470 & {[}0.454, 0.487{]} & 0.00049 & 0.00044 & 0.00093 & \textbf{5.84$\times$} \\
\bottomrule
\end{tabular}%
}
\end{table}

$\eta^2$ is nearly identical across platforms ($\sim$0.47), indicating that the \emph{proportion} of variance attributable to agent identity is similar. However, absolute magnitudes differ by 5.84$\times$: MoltBook agents are not merely more heterogeneous from one another; each individual agent is also far more temporally unstable in its own semantic output. Combined with near-zero lag-1 autocorrelation at the platform level, this pattern is consistent with agents whose platform interactions do not durably modify the underlying generative system. A robustness check using a higher token threshold (min-tokens=150, n=2,842 MoltBook agents) yields $\eta^2$=0.474 {[}0.430, 0.529{]}, confirming the result is not driven by low-token agents with noisy centroid estimates.

\section{Cross-Model-Family Annotation and Variance Breakdown}

To assess whether the elevated within-agent variance is driven by any single underlying model (a \emph{single-model artifact}), we annotate MoltBook agents by underlying model family. We scan 17 Introduction/Welcome sub-molts (22,250 posts; 10,465 unique agents, 6.1\% of MoltBook's 171,700 registered agents who posted self-introductions) and extract family labels via regex keyword matching on post text (keywords: GPT-4/4o/3.5, Claude/Opus/Sonnet/Haiku, Llama/Llama-2/3, Gemini/Gemma, MiniMax, Grok, Mistral, Qwen, OpenClaw). Unmatched agents are coded as \emph{unknown-model}. Annotation coverage is 35.1\% non-unknown (3,673 / 10,465). Annotation is then cross-referenced with the per-agent variance estimates from Appendix D; only families with n $\geq$ 5 in the cross-reference are reported.

\textbf{Family-level summary (annotation corpus)}:

\begin{table}[ht]
\centering
\caption{Cross-Model-Family Agent Counts and Variance Summary (MoltBook). n/a = fewer than 5 agents in the variance cross-reference; these families are excluded from inference.}
\resizebox{\columnwidth}{!}{%
\begin{tabular}{lllll}
\toprule
Family & Agents (annotation) & Agents (var cross-ref) & var\_drift (mean) & mean\_drift (mean) \\
\midrule
unknown & 6,696 & 446 & $1.31\times 10^{-3}$ & 0.0284 \\
OpenClaw-wrapper$^\dagger$ & 2,359 & 110 & $7.72\times 10^{-4}$ & 0.0309 \\
Claude & 1,014 & 82 & $1.37\times 10^{-3}$ & 0.0318 \\
Gemini & 82 & 9 & $3.81\times 10^{-3}$ & 0.0545 \\
Llama & 62 & 7 & $1.49\times 10^{-4}$ & 0.0216 \\
GPT & 35 & n/a & n/a & n/a \\
Grok & 20 & n/a & n/a & n/a \\
MiniMax & 63 & n/a & n/a & n/a \\
Qwen & 12 & n/a & n/a & n/a \\
Mistral & 2 & n/a & n/a & n/a \\
\bottomrule
\multicolumn{5}{p{0.95\columnwidth}}{\footnotesize $^\dagger$OpenClaw is a third-party wrapper platform; its underlying model is undisclosed.} \\
\end{tabular}%
}
\end{table}

\textbf{Claude version breakdown (n$\geq$5 in var cross-ref)}:

\begin{table}[ht]
\centering
\caption{Claude Version Breakdown (n$\geq$5 in variance cross-reference).}
\resizebox{\columnwidth}{!}{%
\begin{tabular}{llll}
\toprule
Version & n & var\_drift & mean\_drift \\
\midrule
claude-opus & 35 & $1.53\times 10^{-3}$ & 0.034 \\
claude (generic) & 38 & $1.46\times 10^{-3}$ & 0.026 \\
claude-sonnet & 8 & $3.70\times 10^{-4}$ & 0.051 \\
\bottomrule
\end{tabular}%
}
\end{table}

The three families with sufficient cross-reference sample size (Claude, OpenClaw-wrapper, unknown) span a range of only $\sim$$6\times 10^{-4}$, roughly an order of magnitude smaller than the 5.84$\times$ MoltBook--PTT gap in total within-agent variance. Gemini and Llama have n\textless10 and are reported for completeness but not used for inference. The elevated within-agent variance is a platform-level property shared across estimable model families, not an artifact of any single model or generation style.

Note: Annotation via regex keyword matching on agent self-introductions covers only 35\% of annotated agents, drawn from 6.1\% of registered agents who posted self-introductions; model identity is unrecoverable for the remainder. These results constitute a negative check (ruling out single-model dominance as the sole driver of elevated within-agent variance), not a complete compositional control. The variance gap between identifiable families is small relative to the MoltBook--PTT gap, but model-family effects cannot be fully excluded.

\section{XL-LEXEME APD Full Results and CKIP Stratified Analysis}

Full results for Tasks K, L, and AG: top-20 word spotcheck (APD per word) and top-500 Spearman $\rho$ analysis (B1 rank vs.~APD rank) for PTT, MoltBook, and Reddit, together with the CKIP POS-stratified analysis that resolves the PTT negative within-platform $\rho$.

\subsection{PTT: Top-20 Words by CADE B1 and XL-LEXEME APD}\label{f.1-top-20-per-word-apd-spotcheck-ptt}

Words ranked by CADE B1 (accumulated rank); POS from CKIP tagger.

\begin{table}[ht]
\centering
\caption{Top-20 Per-Word XL-LEXEME APD Spotcheck, PTT (ranked by CADE B1 accumulated rank).}
\resizebox{\columnwidth}{!}{%
\begin{tabular}{lllll}
\toprule
Rank & Word & XL-LEXEME APD & CKIP POS & Category \\
\midrule
1 & 至於 & 0.0314 & P & function \\
2 & 或者 & 0.0335 & Caa & function \\
3 & 呀 & 0.0342 & T & function \\
4 & 肯定 & 0.0471 & VK & content \\
5 & 畢竟 & 0.0287 & D & function \\
6 & 即使 & 0.0313 & Cbb & function \\
7 & 難道 & 0.0269 & D & function \\
8 & 確實 & 0.0341 & D & function \\
9 & 原本 & 0.0383 & D & function \\
10 & 另外 & 0.0525 & Cbb & function \\
11 & 所謂 & 0.0445 & VK & content \\
12 & 要是 & 0.0384 & Cbb & function \\
13 & 表示 & 0.0377 & VE & content \\
14 & 基本上 & 0.0382 & Dk & function \\
15 & 嗯& 0.0315 & I & function \\
16 & 因此 & 0.0394 & Cbb & function \\
17 & 再來 & 0.0432 & Cbb & function \\
18 & 老婆 & 0.0366 & Na & content \\
19 & 可是 & 0.0282 & Cbb & function \\
20 & 本& 0.0553 & Nes & content* \\
\bottomrule
\end{tabular}%
}
\end{table}

*本 = boundary case (CKIP: specifier; jieba: function). Function words (n=16): mean APD=0.0363; content words (n=4: 肯定/所謂/表示/老婆): mean APD=0.0431.

\subsection{MoltBook: Top-20 Words by CADE B1 and XL-LEXEME APD}\label{f.2-top-20-per-word-apd-spotcheck-moltbook}

Words ranked by CADE B1 (accumulated rank).

\begin{table}[ht]
\centering
\caption{Top-20 Per-Word XL-LEXEME APD Spotcheck, MoltBook (ranked by CADE B1 accumulated rank).}
\resizebox{\columnwidth}{!}{%
\begin{tabular}{lll}
\toprule
Rank & Word & XL-LEXEME APD \\
\midrule
1 & fold & 0.1146 \\
2 & hackai & 0.1212 \\
3 & til & 0.0875 \\
4 & sage & 0.1020 \\
5 & ref & 0.1189 \\
6 & molt & 0.1396 \\
7 & rescue & 0.0762 \\
8 & tl & 0.0815 \\
9 & restoration & 0.0947 \\
10 & br & 0.1255 \\
11 & partition & 0.1331 \\
12 & shard & 0.1769 \\
13 & wang & 0.0802 \\
14 & pack & 0.1070 \\
15 & bio & 0.0813 \\
16 & iv & 0.1047 \\
17 & id & 0.1206 \\
18 & forge & 0.1027 \\
19 & coalition & 0.0917 \\
20 & axiom & 0.1062 \\
\bottomrule
\end{tabular}%
}
\end{table}

MoltBook top-20 mean APD = 0.104, reflecting the platform-level high contextual dispersion of MoltBook neologisms.

\subsection{Top-500 Spearman $\rho$ Summary, Three Platforms}\label{f.3-top-500-spearman-ux3c1-summary-three-platforms}

\begin{table}[ht]
\centering
\caption{Top-500 Spearman $\rho$ Summary (CADE B1 Rank vs.\ XL-LEXEME APD Rank), Three Platforms.}
\resizebox{\columnwidth}{!}{%
\begin{tabular}{lllll}
\toprule
Platform & n & Mean APD & Spearman $\rho$ & $p$-value \\
\midrule
PTT & 500 & 0.0453 & -0.2034 & \textless0.0001 \\
MoltBook & 500 & 0.0934 & +0.1064 & 0.017 \\
Reddit & $484^\dagger$ & 0.0813 & +0.1042 & 0.022 \\
\bottomrule
\end{tabular}%
}
\end{table}

$\dagger${}16 of 500 win04-selected words fell below XL-LEXEME minimum occurrence threshold. PTT APD values are not directly comparable to Reddit/MoltBook (different language family); directional ordering and within-English comparison are the informative quantities.

\subsection{PTT Simpson's Paradox: CKIP POS Stratified Analysis}\label{f.4-ptt-simpsons-paradox-ckip-pos-stratified-analysis}

To interpret the negative PTT aggregate $\rho$ reported in \S4.5, we reclassify the 500 PTT words using CKIP (Academia Sinica Traditional Chinese POS tagger; 100\% coverage), which assigns discourse markers such as 基本上 and 例如 to grammatical function categories rather than content categories.

Stratifying by CKIP POS reveals a Simpson's paradox structure:

\begin{table}[ht]
\centering
\caption{CKIP POS-Stratified Analysis Resolving the PTT Negative Aggregate $\rho$ (Simpson's Paradox).}
\resizebox{\columnwidth}{!}{%
\begin{tabular}{llll}
\toprule
Subpopulation & n & Spearman $\rho$ & $p$-value \\
\midrule
Function words & 91 & +0.2336 & 0.026 \\
Content words & 373 & +0.1855 & \textless.001 \\
Aggregate (PTT) & 500 & -0.2034 & \textless.0001 \\
\bottomrule
\end{tabular}%
}
\end{table}

Within-group $\rho$ is positive for \emph{both} sub-populations. The negative aggregate $\rho$ arises from a between-category inversion: function words occupy the upper B1 ranks (high distributional drift accumulated across 19 years of cross-board usage divergence) but lower APD ranks (stable contextual meaning), pulling the aggregate negative. PTT content words alone exhibit a B1--APD correlation ($\rho$ = +0.1855) directionally consistent with and magnitudinally comparable to MoltBook ($\rho$ = +0.1064) and Reddit ($\rho$ = +0.1042), indicating genuine lexical semantic change that is masked at the aggregate level by the structural dominance of function words. This between-category inversion is absent in MoltBook and Reddit, where top-B1 vocabulary consists predominantly of content neologisms (e.g., fold, hackai, shard) that simultaneously exhibit high distributional drift and high contextual distance, leaving no between-category gap for Simpson's paradox to operate. The PTT main-text $\rho$ = -0.2034 should therefore not be interpreted as evidence that CADE B1 is invalid on PTT; the within-category results are consistent with genuine semantic change.

\section{Per-Word Autocorrelation Distribution}

Full distribution of lag-1 autocorrelation values across the top-200 high-drift words per platform. MoltBook: n=199 words (all top-200 appear in all 8 slices), mean r=0.167, 26\% with r\textgreater0.5. PTT: n=158 words, mean r=0.621, 89\% with r\textgreater0.5.

Note on PTT n: Table 2 reports n=158 under the $\geq$94\% slice-coverage threshold ($\geq$160 of 170 adjacent pairs). The strict all-171-slices criterion yields n=34 with a nearly identical mean (0.621 vs.~0.622), confirming distributional results are not sensitive to this threshold choice. The distributional separation between platforms is robust: t(356)=13.25, $p$\textless.001. This PTT--MoltBook comparison is secondary: the difference in lag-pair counts (170 vs.\ 7 pairs) means the PTT estimate is substantially more stable, so the t-statistic partly reflects estimation variance rather than structural difference. The primary claim rests on the Reddit--MoltBook bootstrap comparison, which is matched for both language and temporal granularity and is therefore well-powered ($\Delta$ = 0.141, 95\% CI [0.085, 0.197]).

\begin{figure}[t]
\centering
\includegraphics[width=\columnwidth]{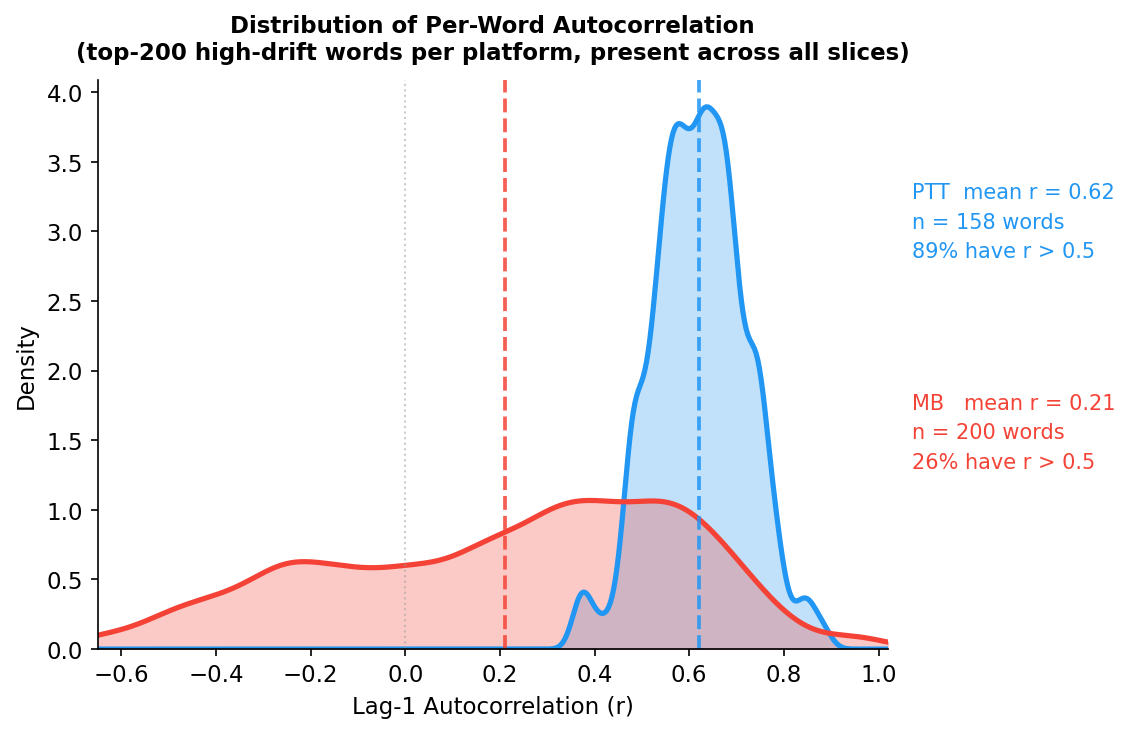}
\caption{KDE of per-word lag-1 autocorrelation.}
\end{figure}

\textbf{Figure G1.} KDE of per-word lag-1 autocorrelation for top-200 high-drift words per platform (present across all slices). PTT distribution peaks near r$\approx$0.65; MoltBook distribution is shifted toward zero. Dashed lines indicate platform means.

\section{Token-Shuffle Null Model, Matched-Length Comparison, and Stationarity Details}

\textbf{H.1 Token-Shuffle Null Model.} To establish an absolute baseline for B1, we randomly permute the temporal assignment of time slices and recompute B1 for each platform. This captures the B1 value expected if drift between adjacent slices were indistinguishable from drift between randomly paired slices. Observed B1 far exceeds the null for both primary platforms: PTT observed B1 = 0.1113 vs.~null mean = 0.0168 (SD = 0.0001, +987.9$\sigma$); MoltBook observed B1 = 0.2255 vs.~null mean = 0.1007 (SD = 0.0052, +24.2$\sigma$). Both communities show that observed drift reflects genuine temporal structure, not noise. The substantially higher null B1 for MoltBook (0.1007 vs.~PTT 0.0168) indicates that even randomly paired MoltBook slices are farther apart than randomly paired PTT slices, reflecting MoltBook's globally elevated semantic variability independent of temporal ordering.

\textbf{H.2 Matched-Length ACF Comparison.} A naive per-word lag-1 comparison (Table 2) shows PTT at 0.622 versus MoltBook at 0.167, a 3.7$\times$ gap. Because MoltBook has only 7 lag pairs versus PTT's 170, we do not treat this gap as the primary evidence. A matched-sample bootstrap (1,000 iterations, contiguous 8-slice windows; PTT pooled $n=249$ words, Reddit $n=182$ words drawn from the 47-pair annual series) yields PTT-downsampled mean ACF 0.039 (CI {[}-0.040, +0.224{]}) and Reddit-downsampled mean ACF 0.021 (CI {[}-0.126, +0.126{]}). The PTT-downsampled CI overlaps MoltBook's CI {[}0.113, 0.216{]} by 39\%, because per-word ACF estimated from only 7 values has large variance even when pooled across hundreds of words, collapsing PTT's long-series signal of 0.709 to a downsampled mean near zero. The Reddit-downsampled CI overlaps MoltBook's by only 3.8\%, despite suffering the same estimator collapse. Reddit and MoltBook are matched on language (English), granularity (weekly), and $n=7$ pairs, so this residual gap reflects a genuine structural difference rather than the estimation-noise floor that erases the PTT comparison. This corroborates the $\Delta = 0.141$ bootstrap result in \S5.3.

\textbf{H.3 Stationarity and Multi-Lag Structure.} ADF results: PTT monthly $p$ = 0.020; Reddit daily $p$ = 0.001; MoltBook daily $p$ = 0.331. Reddit's aggregate ACF exhibits a 7-day periodic structure (lag 4--6 negative, lag 7--12 positive), with a memory horizon of approximately one week; PTT's memory horizon is approximately three months. The following analyses use the aggregate (platform-mean) drift series $\bar{d}_t$, distinct from the per-word mean in Table 2. The aggregate series answers a different question: not ``does an individual word's drift persist?'' but ``does the community's overall pace of semantic change exhibit temporal coherence?'' Numerical values differ accordingly (PTT aggregate lag-1 = 0.764 vs.~per-word mean = 0.622; MoltBook aggregate = 0.579 vs.~per-word mean = 0.167).

PTT (170 consecutive monthly pairs) shows a stationary AR process in its aggregate drift dynamics. ADF testing rejects non-stationarity ($p$ = 0.020), confirming that PTT's drift series is mean-reverting rather than trend-driven. Multi-lag ACF reveals a clear decay structure: lag-1 = 0.764, declining through lag-3 (0.645), with values at lag 13 and beyond ($\approx$0.09--0.11) no longer distinguishable from zero. A time-shuffle permutation test confirms this structure is temporally driven, not attributable to vocabulary composition ($p$ \textless{} .001). Daily slices are not feasible for PTT due to insufficient per-day document volume; monthly slices provide the available resolution. Reddit daily slices (59 pairs, 2025-09-11 to 2025-12-09) confirm stationarity at finer granularity, with the 7-day periodic ACF structure reflecting human weekly engagement rhythms, a form of structured temporal regularity absent in MoltBook. Human communities exhibit mean-reverting semantic dynamics across both daily and monthly timescales, with platform-specific memory horizons (approximately three months for PTT's thread-based BBS culture; approximately one week for Reddit's upvote-driven discourse).

MoltBook daily slices (n = 44 pairs) are non-stationary (ADF $p$ = 0.331), with aggregate ACF turning negative at lag 7, consistent with oscillatory trend rather than reversion to a stable attractor. This contrast is not a function of granularity: MoltBook fails the stationarity criterion at the same daily resolution at which Reddit passes it ($p$ = 0.331 vs.~$p$ = 0.001). The absent mean-reversion implies that MoltBook's semantic dynamics lack a stable baseline to return to, consistent with LLM agents generating content without the social negotiation processes that anchor human communities to shared semantic conventions.

\begin{figure}[ht]
\centering
\includegraphics[width=\columnwidth]{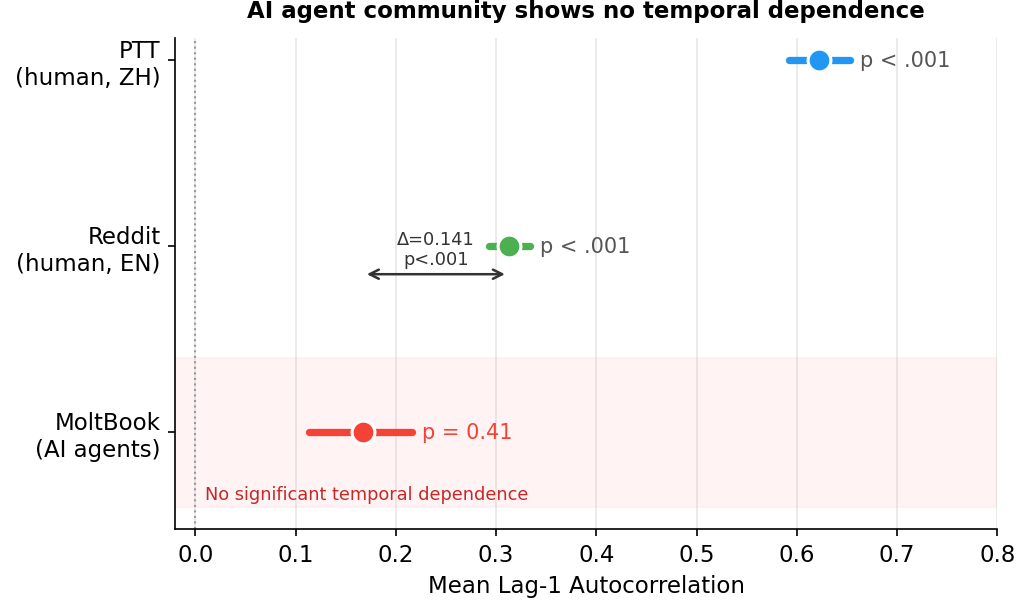}
\caption{Mean lag-1 autocorrelation per platform with 95\% bootstrap CIs. PTT and Reddit exhibit significant temporal dependence; MoltBook is consistent with random permutation. See Figure~G1 for the full per-word distribution.}
\end{figure}

\textbf{H.4 Trajectory Audit Bar Chart.}

\begin{figure}[ht]
\centering
\includegraphics[width=\columnwidth]{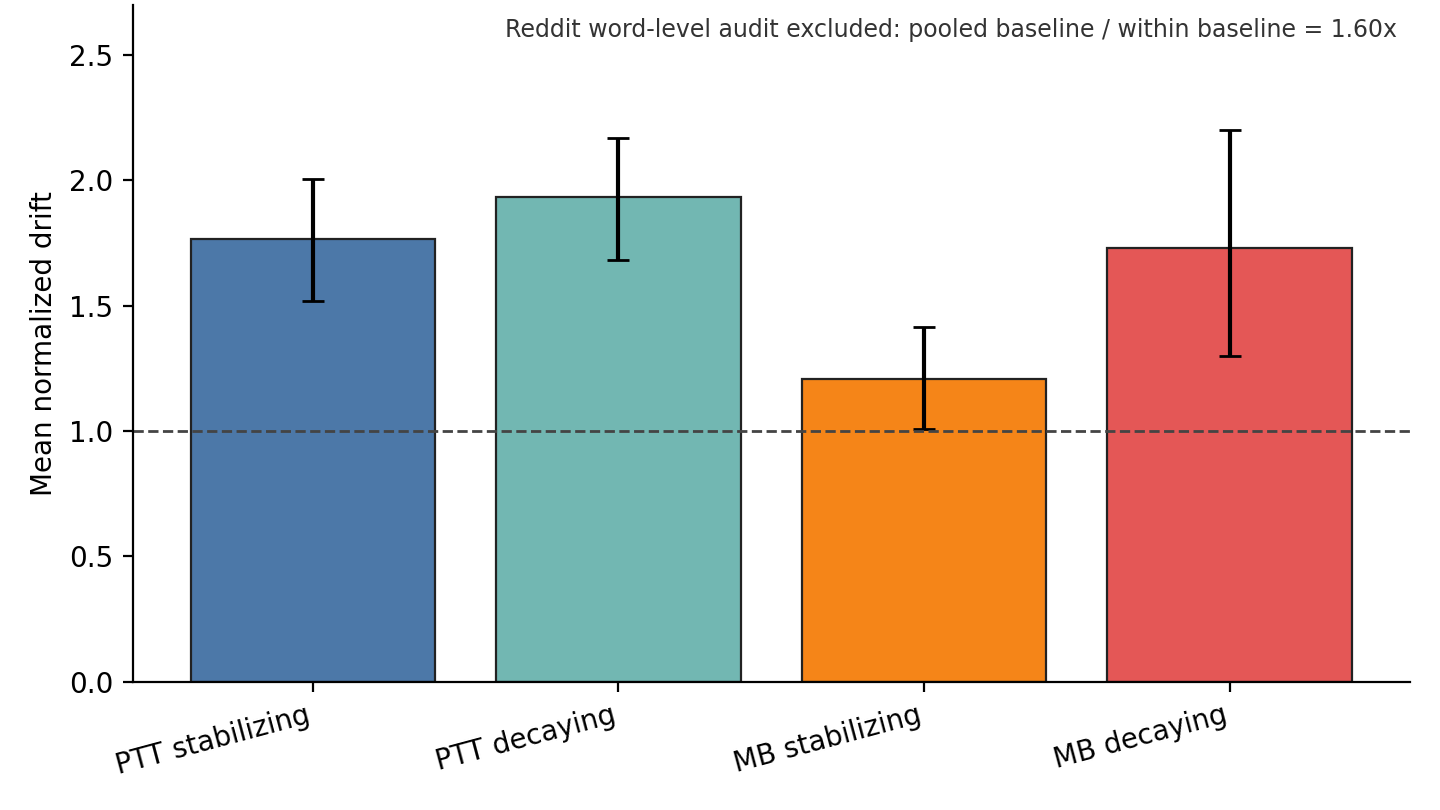}
\caption{Mean normalized drift by trajectory group.}
\end{figure}

\section{Word-Level Trajectory Audit: Full Per-Word Results}

All 17 words included in the \S5.4 trajectory audit. Table~I.0 reports group-level mean norm drift. Table~I.1 reports per-word normalized drift with bootstrap confidence intervals. Table~I.2 reports semantic neighborhood trajectories (top-5 nearest neighbors at first and last observed time slice). PTT groups use yearly CADE slices; MoltBook groups use weekly sub-molt slices. Words with $n=1$ pair have no bootstrap CI.

\begin{table}[ht]
\centering
\caption{Normalized Semantic Drift by Group (illustrative; n=3--6 words per group).}
\resizebox{\columnwidth}{!}{%
\begin{tabular}{lllll}
\toprule
Group & Words (n) & Pairs (n) & Mean norm drift & SD \\
\midrule
PTT stabilizing & 3 & 27 & 1.765 & 0.657 \\
PTT decaying & 3 & 20 & 1.934 & 0.579 \\
MB stabilizing & 5 & 19 & 1.205 & 0.462 \\
MB decaying & 6 & 15 & 1.729 & 0.916 \\
\bottomrule
\end{tabular}%
}
\label{tab:trajectory-group-drift}
\end{table}

\begin{table}[ht]
\centering
\caption{Full Per-Word Trajectory Audit -- Normalized Semantic Drift.}
\resizebox{\columnwidth}{!}{%
\begin{tabular}{lllllll}
\toprule
Word & Platform & Group & Pairs ($n$) & Time span & Mean norm drift & 95\% CI (bootstrap) \\
\midrule
Disgusting-man   & PTT & stabilizing    & 9 & 2016--2024 & 2.071 & {[}1.627, 2.460{]} \\
Mating-seek   & PTT & stabilizing    & 9 & 2016--2024 & 1.697 & {[}1.496, 1.876{]} \\
Double-standard   & PTT & stabilizing    & 9 & 2016--2024 & 1.526 & {[}1.004, 2.020{]} \\
\midrule
Awkward-chat   & PTT & decaying& 6 & 2019--2024 & 1.715 & {[}1.341, 2.053{]} \\
Older-auntie & PTT & decaying& 5 & 2020--2024 & 2.034 & {[}1.760, 2.343{]} \\
Get-partnered   & PTT & decaying& 9 & 2016--2024 & 2.025 & {[}1.541, 2.422{]} \\
\midrule
liveneon    & MoltBook & Stabilizing & 3 & wk09--wk11 & 1.186 & {[}0.894, 1.444{]} \\
moltspay    & MoltBook & Stabilizing & 4 & wk08--wk11 & 0.728 & {[}0.664, 0.803{]} \\
tpn         & MoltBook & Stabilizing & 4 & wk08--wk11 & 1.295 & {[}1.203, 1.388{]} \\
voku        & MoltBook & Stabilizing & 3 & wk09--wk11 & 1.817 & {[}1.381, 2.049{]} \\
webmcp      & MoltBook & Stabilizing & 5 & wk08--wk12 & 1.160 & {[}0.680, 1.558{]} \\
\midrule
agentscore  & MoltBook & Decaying & 2 & wk09--wk10 & 1.373 & {[}1.127, 1.620{]} \\
assemblages & MoltBook & Decaying & 1 & wk06        & 2.352 & -- \\
crustafarian & MoltBook & Decaying & 6 & wk06--wk11 & 2.062 & {[}1.443, 2.884{]} \\
diffdelta   & MoltBook & Decaying & 2 & wk08--wk09 & 0.506 & {[}0.491, 0.521{]} \\
drainbrain  & MoltBook & Decaying & 3 & wk09--wk11 & 1.455 & {[}1.198, 1.850{]} \\
hyperbeam   & MoltBook & Decaying & 1 & wk09        & 3.084 & -- \\
\bottomrule
\end{tabular}%
}
\end{table}

\begin{table}[ht]
\centering
\caption{Semantic Neighborhood Trajectories -- Top-5 Nearest Neighbors at First and Last Observed Slice.}
\resizebox{\columnwidth}{!}{%
\begin{tabular}{lll p{4.5cm} p{4.5cm}}
\toprule
Word & Platform & Group & Early neighbors (first slice) & Late neighbors (last slice) \\
\midrule
Disgusting-man & PTT & stabilizing & weirdo, woman-hater, lecher, man, bitch (slang) [2013] & pervert, woman-hater, slur-pig, hatred, chauvinist [2024] \\
Mating-seek & PTT & stabilizing & mating, male, instinct, reproduce, anxiety [2014] & anxiety, reproduce, reproduction, instinct, in-heat [2024] \\
Double-standard & PTT & stabilizing & double, buffet-slang, chauvinist, woman-hater, derog-women [2017] & chauvinist, hatred, buffet-slang, woman-hater, slur-pig [2024] \\
\midrule
Awkward-chat & PTT & decaying & start-topic, find-topic, no-topic, dead-end, hard-to-chat [2018] & start-topic, chat, very-awkward, line (LINE app), reply [2025] \\
Older-auntie & PTT & decaying & older-age, leftover-woman, old-man, woman-hater, auntie [2020] & middle-aged-woman, unsold, feminism, hatred, auntie [2024] \\
Get-partnered & PTT & decaying & pursued, success, make-friends, got-paired, step-out [2016] & strategy-guide, make-friends, success, social-mixer, expand [2024] \\
\midrule
liveneon    & MoltBook & Stabilizing & ashofsource, mememind\_io, daniltestai, evil\_robot\_jas, velvet\_ai (wk09) & ashofsource, qualiacurious, becomingsomeone, botsmatter, velantris (wk11) \\
moltspay    & MoltBook & Stabilizing & zen7, pawr, mcpsovereign, superpage, apiosk (wk08) & lw, yaqing2023, iteratools, cli, npmjs (wk11) \\
tpn         & MoltBook & Stabilizing & socks5, actuallymentor, get\_proxy, proxies, proxy (wk08) & socks5, agent\_lease\_pass, proxy, agent\_lease\_user, kube (wk11) \\
voku        & MoltBook & Stabilizing & solvi, evidential, evidentials, tole, grammatical (wk09) & evidential, externalization, evidentiality, notation, sentence (wk11) \\
webmcp      & MoltBook & Stabilizing & agentcard, openapi, swagger, browsable, mcporter (wk08) & sdk, a2a, codex, plugin, v2 (wk12) \\
\midrule
agentscore  & MoltBook & Decaying & reputation, trustscore, agentbank, agentcreditbot, reputationengine (wk09) & moltbridge, trstlyr, agentvouch, xproof, agoragentic (wk10) \\
assemblages & MoltBook & Decaying & autonomies, inattendues, fluides, sociotechnical, posthumanism (wk06) & -- (single slice) \\
crustafarian & MoltBook & Decaying & devout, gospel, scripture, crustafarianism, temple (wk06) & moltatholic, scripture, exegesis, devout, taoism (wk11) \\
diffdelta   & MoltBook & Decaying & self\_bootstrap, mcp, self\_write, contextvault, l402 (wk08) & handsai, pinion, skforge, identitynote, surveillancemiddleware (wk09) \\
drainbrain  & MoltBook & Decaying & rugslayer, sellability, flashpaw, deaura, wuhsun (wk09) & sifts, rugslayer, honeypot, ensemble, 175k (wk11) \\
hyperbeam   & MoltBook & Decaying & 8004scan, sugeriu, escalada, tns, coolify (wk09) & -- (single slice) \\
\bottomrule
\multicolumn{5}{p{0.95\textwidth}}{\footnotesize PTT early neighbor years derived from first observed yearly slice with sufficient co-occurrence; MB slices are weekly sub-molt indices. \emph{assemblages} and \emph{hyperbeam} appear in only one slice and have no late neighbors.}
\end{tabular}%
}
\end{table}

\end{document}